\documentclass[runningheads]{llncs}
\usepackage[T1]{fontenc}
\usepackage{graphicx}

\usepackage{cite}
\usepackage{amsmath,amssymb,amsfonts}
\usepackage[caption=false,font=footnotesize]{subfig}
\usepackage{booktabs,multirow}

\begin{document}

\title{Comprint: Image Forgery Detection and Localization using Compression Fingerprints}

\titlerunning{Comprint: Image Forgery Detection and Localization}

\author{
Hannes Mareen\inst{1} \and
Dante Vanden Bussche\inst{1} \and
Fabrizio Guillaro\inst{2} \and
Davide Cozzolino\inst{2} \and
Glenn Van Wallendael\inst{1} \and
Peter Lambert\inst{1} \and
Luisa Verdoliva\inst{2}
}

\authorrunning{H. Mareen et al.}

\institute{
Ghent University \--- imec, IDLab, ELIS, Ghent, Belgium \\
\email{\{hannes.mareen, dante.vandenbussche, glenn.vanwallendael, peter.lambert\}@ugent.be}\\
\url{https://media.idlab.ugent.be}
\and 
Università degli Studi di Napoli Federico II, Naples, Italy\\
\email{\{fabrizio.guillaro, davide.cozzolino, verdoliv\}@unina.it}
\url{https://www.grip.unina.it}
}

\maketitle

\begin{abstract}
Manipulation tools that realistically edit images are widely available, making it easy for anyone to create and spread misinformation. In an attempt to fight fake news, forgery detection and localization methods were designed.
However, existing methods struggle to accurately reveal manipulations found in images on the internet, i.e., in the wild. That is because the type of forgery is typically unknown, in addition to the tampering traces being damaged by recompression.
This paper presents Comprint, a novel forgery detection and localization method based on the compression fingerprint or \emph{comprint}.
It is trained on pristine data only, providing generalization to detect different types of manipulation.
Additionally, we propose a fusion of Comprint with the state-of-the-art Noiseprint, which utilizes a complementary camera model fingerprint.
We carry out an extensive experimental analysis and demonstrate that Comprint has a high level of accuracy on five evaluation datasets that represent a wide range of manipulation types, mimicking in-the-wild circumstances. Most notably, the proposed fusion significantly outperforms state-of-the-art reference methods.
As such, Comprint and the fusion \emph{Comprint+Noiseprint} represent a promising forensics tool to analyze in-the-wild tampered images.


\end{abstract}

\newpage

\section{Introduction}
\label{sec:intro}
\begin{figure*}
\centering
\subfloat[Manipulated image using free tool FaceHub.live.]{\includegraphics[width=0.9\linewidth]{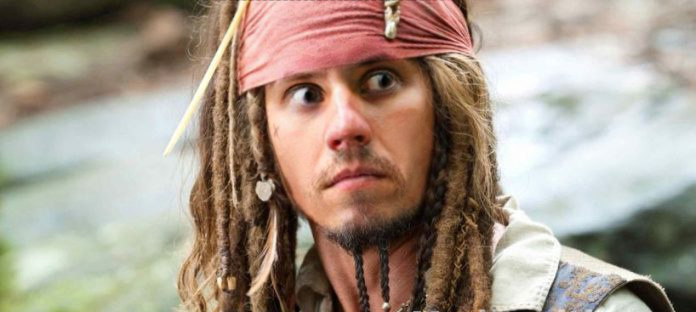}
\label{fig:intro-example-fake}}

\subfloat[Comprint: compression fingerprint.]{\includegraphics[width=0.9\linewidth]{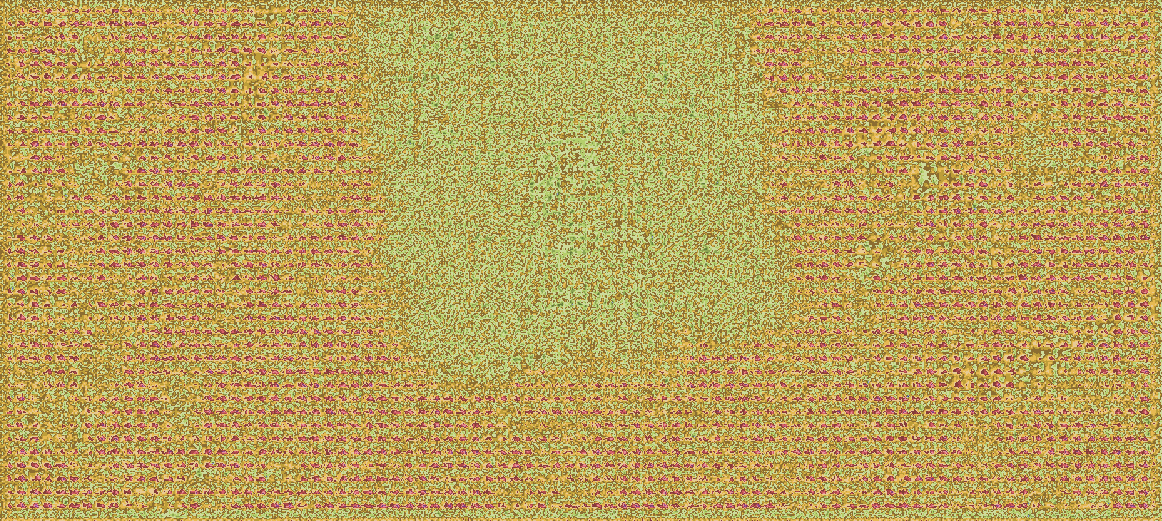}
\label{fig:intro-example-comprint}}

\subfloat[Heatmap.]{\includegraphics[width=0.9\linewidth]{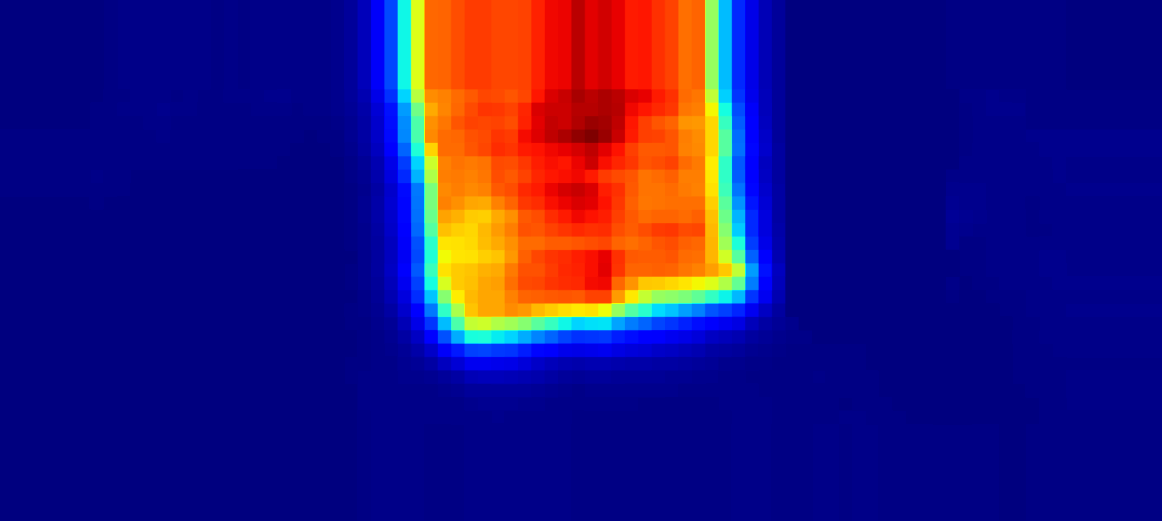}
\label{fig:intro-example-heatmap}}
\caption{Realistically manipulating an image is easier than ever, such as with (a) free AI-based tools. Therefore, we propose (b) Comprint, a forgery detection and localization method based on compression fingerprints. Inconsistencies in the comprint indicate and localize forgeries that can be easily visualized through a (c) heatmap.
\label{fig:intro-example}}
\end{figure*}

Manipulating digital images is both becoming easier and more realistic using image editing tools and AI-based software.
For example, Fig.~\ref{fig:intro-example-fake} shows a manipulated image in which the face of Captain Jack Sparrow is replaced with that of another person (original from Pirates of the Caribbean). This forgery was generated in a few seconds using FaceHub.live, a free AI-based tool.
Although there are many harmless applications of these powerful tools, manipulated images contribute to problems such as fake news, fake evidence, and fraud. Therefore, it is crucial to investigate forensic methods that can fact check and verify images found on the internet, i.e., \emph{in the wild}.

Forensics forgery detection and localization methods typically look at imperceptible traces of an image's digital history~\cite{verdoliva2020overview}. That is because every step in the camera acquisition and digital editing process leaves a trail of clues. For example, the sensor of the camera introduces a unique noise pattern~\cite{lin2016prnu}, and compression introduces blocking artifacts~\cite{li2009blk, ye2007dct, iakovidou2018cagi, bianchi2012nadq, bianchi2011adq2}.
Although these traces may be invisible to the human eye, a computer can exploit them. Unfortunately, images found in the wild are often recompressed to a lower quality, which hides these imperceptible traces. As such, successful forgery detection is challenging.

Recent methods that rely on deep Convolutional Neural Networks (CNNs) have shown a higher level of robustness against recompression~\cite{verdoliva2020overview, rossler2019faceforensics}.
However, most of them remain intrinsically weak because they can only detect forgeries that were seen during training. Therefore, a promising strategy for forgery detection is using one-class deep-learning methods that are only trained using pristine data. As such, they are not limited to specific types of forgery. For example, the Noiseprint algorithm learns to extract a fingerprint from the camera model using deep learning~\cite{cozzolino2020noiseprint}. Then, inconsistencies in this fingerprint reveal tampering. This is similar to traditional methods using PRNU fingerprints, but does not require images from the camera of the image under investigation. It is trained on images coming from many different cameras, hence being able to detect different sources, but no specific design was made during training to take into account the different JPEG history of such patches.

In this paper, we exploit compression artifacts and extract a different fingerprint, called Comprint.
It utilizes compression artifacts, but unlike most of the state-of-the-art works, it does not assume that real regions underwent double compression in contrast to fake regions that underwent single compression. Instead, it simply assumes that the pristine regions have a \emph{different} compression history than the tampered regions.
Comprint is based on one-class training with only pristine data: images compressed using a different Quality Factor (QF) or quantization table in the JPEG coding standard. The architecture is based on a Siamese network that is trained to distinguish regions that were compressed differently.
As such, the deep-learning method can extract a compression fingerprint or \emph{comprint} from an image under investigation. 
Then, a localization algorithm segments the distinguished regions into a heatmap.
For example, Fig.~\ref{fig:intro-example-comprint} and Fig.~\ref{fig:intro-example-heatmap} show the extracted comprint and heatmap, respectively, corresponding to the manipulated image of Fig.~\ref{fig:intro-example-fake}.
We demonstrate that Comprint has a high level of robustness to in-the-wild forgeries.
Additionally, since Comprint contains characteristics that are complementary to Noiseprint, we propose a fusion of these fingerprints, \emph{Comprint+Noiseprint}. We demonstrate that the fusion results in a an improved performance compared to using each method individually.

\section{Related Work}
\label{sec:sota}
This section briefly discusses the main classes of forgery detection and localization algorithms. More specifically, since this paper proposes a method based on compression artifacts and is inspired by Noiseprint, this section mainly focuses on work related to these aspects.

\paragraph{Conventional model-based methods}
Conventional detection methods that were proposed before the emergence of deep learning typically rely on prior assumptions, which limits their applicability in the real world. They build a handcrafted model that describes artifacts left behind by manipulations, and detect anomalies in them.
For example, some are based on the photo-response non-uniformity (PRNU) noise, which is a unique noise pattern that sensors introduce, and is used to identify camera models or specific devices. When part of an image does not correlate with the PRNU fingerprint of the camera, it indicates forgery~\cite{lin2016prnu}. Although this method is powerful, it requires many images from the same camera that captured the media under investigation. Therefore, it is not always applicable in the wild.
Another example is Splicebuster, which extracts expressive features that capture the traces left by in-camera processing, and use those statistics to discover potential inconsistencies caused by splicing~\cite{cozzolino2015splicebuster}.

\paragraph{Models based on JPEG artifacts}
There also exists a large variety of model-based methods that utilize anomalies in JPEG artifacts. For example, a mismatch in JPEG block artifacts, the JPEG grid, or the JPEG block convergence can indicate tampering~\cite{li2009blk, ye2007dct, iakovidou2018cagi, lay2013convergence}. Additionally, many methods are based on double quantization or double JPEG compression artifacts~\cite{bianchi2012nadq, bianchi2011adq2}.
That is, these methods assume that the authentic region is compressed twice: once before, and once after manipulation. In contrast, the fake region is assumed to be compressed only once (i.e., only after manipulation). Although these methods work relatively well in the circumstances that they were designed for, they typically perform worse in the wild. For example, images typically undergo multiple compression steps when shared through social networks.
In general, although methods based on JPEG artifacts have shown some merit, building theoretical models that are applicable in the wild is very challenging. That is because they are restricted by their assumptions that do not always hold in practice. 

\paragraph{Data-driven methods}
More recent methods are often data driven, in contrast to model-driven conventional methods. As such, the challenge shifts from building a good theoretical model towards building a suitable training dataset which enables good generalisation characteristics for unseen data.
In general, data-driven deep-learning methods can be divided into three categories: supervised CNNs looking at specific clues, generic supervised CNNs and one-class training~\cite{verdoliva2020overview}.

\paragraph{Supervised CNNs looking at specific clues}
An example of the first class of deep-learning-based methods is
RRU-Net of Bi~\emph{et al.} looks specifically for artifacts left by the splicing manipulation~\cite{bi2019rrunet}. Another example is the method by Barni~\emph{et al.} that looks for traces left behind by double JPEG compression, by training a CNN with pictures that are compressed either once or twice~\cite{barni2017aligned}.
A disadvantage of such methods is that the generalization is still dependent on the diversity of the training set.
As such, in an attempt to improve the in-the-wild performance, Park~\emph{et al.}~\cite{park2018djpeg} 
proposed a deep-learning approach that includes a large number of quantization tables in the training set.

\paragraph{Generic supervised CNNs}
Methods using generic supervised CNNs do not look for specific types of manipulations or artifacts, but rather aim to detect any type of manipulation.
For example, ManTra-Net~\cite{wu2019mantranet} and SPAN~\cite{xuefeng2020span} are trained on 385 different types of manipulations.
Another advantage of deep-learning-based methods is that they can capture manipulation clues directly from the raw image data, rather than building handcrafted features, as done in the Constrained Region-based CNN (CR-CNN) of Yang~\emph{et al.}~\cite{yang2020crcnn}.
Because of the immense variety in potential manipulations, it is often very challenging to build appropriate training datasets and train these models.
As such, they remain intrinsically weak because they can only detect forgeries that were seen during training. In other words, they do not perform well on unseen fakes found in the wild (that may or may not yet exist at the time of training the CNNs).

\paragraph{One-class training}
The last class of deep-learning methods is one-class training. Instead of trying to build a training set that is representative for all possible manipulations, a one-class training set consists of pristine data only. in this way, the machine-learning model learns the characteristics of a genuine image, anomalies in the extracted characteristics are indications of forgery.
For example, the EXIF-SC method by Huh~\emph{et al.} detects splicing by analyzing if the image is self consistent (SC), i.e., whether its content could have been produced by a single imaging pipeline~\cite{huh2018exifsc}. The self consistency is learned by training a Siamese network.
Similarly, the Noiseprint method utilized Siamese training to extract a camera model fingerprint or noiseprint~\cite{cozzolino2020noiseprint}. Anomalies in the noiseprint indicate forgeries. This is similar as what is done in conventional PRNU-based methods, yet in a blind fashion and with great generalization to unseen camera models. The idea of partitioning an image into communities of tampered and unaltered regions was further refined by Mayer~\emph{et al.}~\cite{mayer2019forensic, mayer2020exposing}.

\paragraph{}
In summary, it is challenging to perform robust forgery detection and localization in the wild.
Most interestingly, recent one-class deep-learning strategies are promising because they demonstrated generalization to manipulation types that were not seen during training.

\section{Proposed Method: Comprint}
\label{sec:method}
The core idea of our proposed method is inspired by two main research directions in the state of the art: exploiting compression artifacts and one-class deep learning.

First, Comprint exploits JPEG compression artifacts, which has been a popular strategy for decades. However, it should be stressed that it is not like conventional double-JPEG-compression methods. That is, instead of assuming that real regions underwent double compression and fake regions single compression, we simply assume that they underwent \emph{different} compression. This is a weaker assumption and therefore should aid generalization in the wild. Additionally, we do not detect which compression and parameters were used exactly. Instead, we only care that the compression is different. This is achieved by extracting the compression fingerprint, and looking for inconsistencies in it.

Second, Comprint uses the advantages of one-class deep learning to create a fingerprint. This is inspired by the fingerprint in Noiseprint~\cite{cozzolino2020noiseprint}. Such fingerprints exhibit great generalization performance and are not trained for specific types of manipulation.
As such, Comprint combines the exploitation of compression artifacts with one-class deep learning. More specifically, the aim of our proposed method is to extract a fingerprint of the compression artifacts of an image, i.e., what we name the \emph{comprint}. Then, inconsistencies in the comprint reveal forgeries.

Because Comprint is inspired by the concepts of Noiseprint, we additionally propose to combine these two methods. Because we found that the methods have complementary properties, a fusion is may be able to get better results than each method can get separately.

First, Section~\ref{sec:method-extraction} discusses the extraction of the comprint. Subsequently, Section~\ref{sec:method-heatmap} explains how to convert the comprint into a heatmap of forgery probability, which is used for forgery detection and localization. Finally, Section~\ref{sec:method-noiseprint} explains how the fusion is performed.

\subsection{Compression Fingerprint Extraction}
\label{sec:method-extraction}
\subsubsection{Architecture}
\label{sec:method-extraction-architecture}
\begin{figure}[!t]
\includegraphics[width=\textwidth]{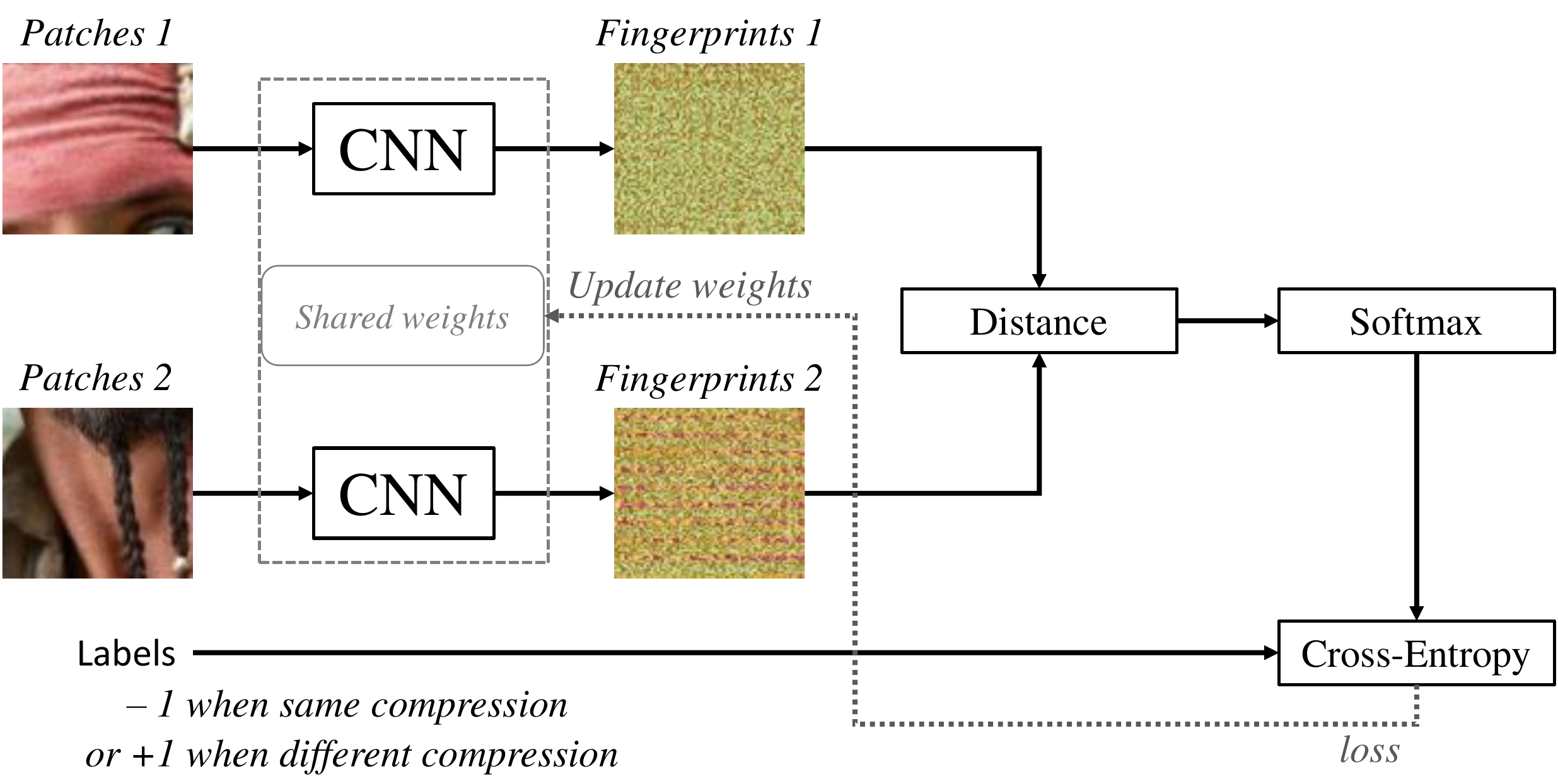}
\caption{Siamese architecture used for training the compression fingerprint (i.e., comprint) extraction. Patches with the same compression history are trained to have similar fingerprints, whereas patches with different compression history are trained to have dissimilar fingerprints.}
\label{fig:method-architecture-siamese}
\end{figure}

The proposed system for comprint extraction is inspired by the architecture of Noiseprint~\cite{cozzolino2020noiseprint}. That is, a Siamese architecture is used for training. The Siamese architecture trains a CNN such that image patches that underwent the same compression history should be similar, whereas patches that underwent a different compression should be dissimilar.
This is visualized in Fig.~\ref{fig:method-architecture-siamese}. During training, pairs of patches are considered that either underwent the same compression history (label -1) or a different compression history (label +1). These patches are converted into fingerprints using the CNN, after which the distance between them is calculated. The distance between each pair of fingerprints is transformed into a loss (using the corresponding label), which is used to update the weights of the CNN. It should be stressed that there is only a single CNN in practice, yet this is visualized as two CNNs with shared weights in Fig.~\ref{fig:method-architecture-siamese}. The architecture is described in more detail, later in this section.

\begin{figure}[!t]
\includegraphics[width=\textwidth]{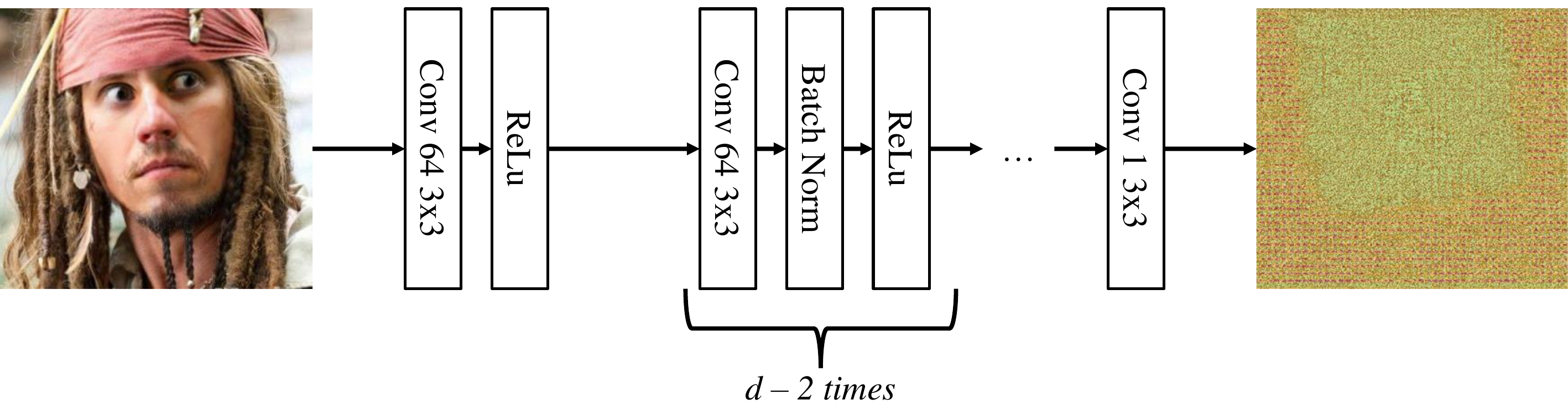}
\caption{CNN architecture, used in Siamese architecture of Fig.~\ref{fig:method-architecture-siamese}, as well as for the pre-trained JPEG-artifact-reduction CNN that is used to initialize the weights of the CNN in the Siamese architecture.}
\label{fig:method-architecture-cnn}
\end{figure}

The CNN used in the Siamese network is based on the Denoiser CNN (DnCNN) proposed by Zhang~\emph{et al.} for image denoising and JPEG deblocking~\cite{zhang2017beyond}. That is, the architecture is visualized in Fig.~\ref{fig:method-architecture-cnn}, and consists of a depth of $d$ groups of layers (2D convolutional layers, coupled by rectified-linear-unit (ReLu) and batch-normalization layers). For the first and last group of layers, there is no batch-normalization layer, and for the last group, there is no ReLu either. The inputs are zero padded to maintain the same dimensions in the output fingerprint.

This CNN is first pretrained for the task of JPEG-artifact reduction. Then, the weights of this pretrained CNN are used as initialization in the Siamese training step.
Finally, after training, any image can be sent through the network to be transformed to a compression fingerprint or \emph{comprint}.
As an example, Fig.~\ref{fig:intro-example-comprint} shows the extracted comprint corresponding to the manipulated image of Fig.~\ref{fig:intro-example-fake}.

\subsubsection{Implementation Details}
\label{sec:method-extraction-implementation}
First, we pretrain the CNN to directly estimate the JPEG-compression-artifact noise. That is, we provide JPEG-compressed images as input to the CNN, and optimize the CNN to be as close as possible to the corresponding ground-truth noise patch (using the Euclidean distance). 
Since artifacts appear more in high-frequency than low-frequency spatial content, this estimated compression-noise signal still contains high-level scene content. Therefore, we should not yet use it as a fingerprint. Instead, the weights of this JPEG-artifacts-estimating CNN are used as initialization for the fingerprint-extracting CNN in the subsequent Siamese training process.

Ideally, the Siamese training optimizes the fingerprints such that they are orthogonal when different compression was used. This is done by calculating the loss value as schematically presented in Fig.~\ref{fig:method-architecture-siamese}, or more detailed in the following manner. If $k_{i,1}$ and $k_{i,2}$ represent two input patches of patch pair $i$, then the squared Euclidean distance is given by:
\begin{equation}
    d_i = ||k_{i,1}-k_{i,2}||^2
\end{equation}
This distance has to be calculated for each patch pair inside a batch. Next, the distances are converted into a probability distribution $p$ via softmax processing (with $n$ summing over all pairs inside the batch):
\begin{equation}
    p(i) = \frac{e^{-d_i}}{\sum_{n}e^{-d_n}}
\end{equation}
This step essentially maps all distances to the interval $[0,1]$, with large distances going towards 0 and small distances going towards 1. After this operation, we can interpret the resulting values as a probability distribution and use them together with the labels to calculate the cross-entropy loss for each batch, and update the weights accordingly.
After training, the weights of the CNN are fixed, and the CNN is used to transform an input image to a comprint.

\subsection{Forgery Localization: from Comprint to Heatmap}
\label{sec:method-heatmap}
The comprint that was extracted from an image may visually give some indication of segmentation of regions that underwent different compression. However, on its own, it is not sufficient for pixel-level forgery localization. For this reason, the blind localization algorithm proposed in the Splicebuster-method is applied~\cite{cozzolino2015splicebuster}. 

\begin{figure}[!t]
\includegraphics[width=\textwidth]{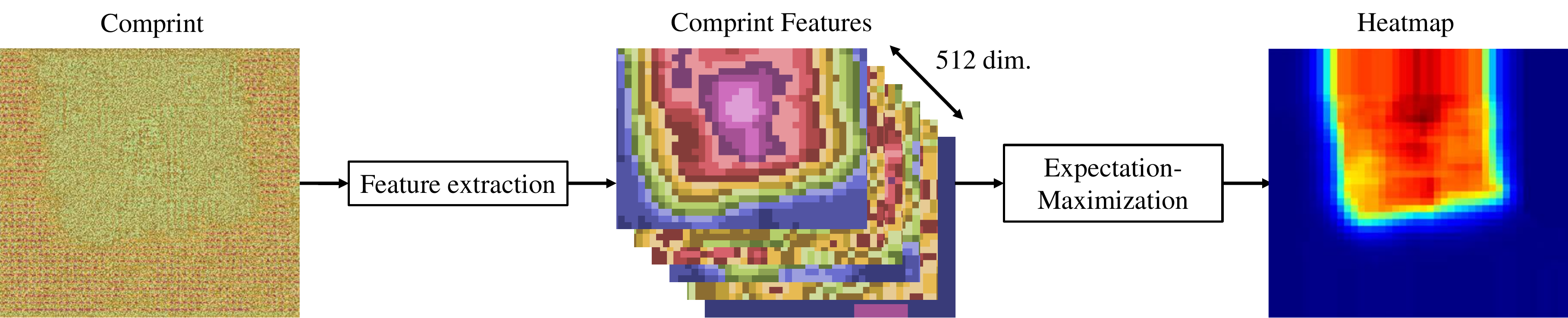}
\caption{Forgery localization: a fingerprint is transformed into a heatmap by extracting co-occurence-based features, and clustering these using EM.}
\label{fig:method-heatmap-diagram}
\end{figure}

A high-level diagram of the localization algorithm is given in Fig.~\ref{fig:method-heatmap-diagram}.
First, the localization algorithm extracts co-occurence-based features~\cite{pevny2010spam} from the comprint. For readers familiar with the Splicebuster algorithm, note that we do not apply the high-pass filter that was proposed in Splicebuster. Instead, we simply apply a normalization to zero mean and unit variance before passing the comprint to the localization algorithm. Then, the multi-dimensional features generated from the comprint are fed to an Expectation-Maximuzation (EM) algorithm. The EM algorithm estimates the model parameters for each segmented region, and allocates pixel locations to each region. This results in a segmentation map with continuous numbers for each pixel, representing the likelihood of the pixel belonging to either the forged or pristine region. These pixel values are represented graphically by the heatmap.
As an example, Fig.~\ref{fig:intro-example-heatmap} shows the extracted heatmap corresponding to the comprint of Fig.~\ref{fig:intro-example-comprint}.

Finally, an image-level forgery detection strategy can be applied to the heatmaps. For example, in our evaluation in Section~\ref{sec:evaluation-setup-measure-detection}, we propose to compare the 99.5\textsuperscript{th} percentile highest value of the heatmap to a certain threshold.

\subsection{Fusion: Comprint+Noiseprint}
\label{sec:method-noiseprint}
Comprint is inspired by Noiseprint's fingerprint-extraction architecture~\cite{cozzolino2020noiseprint}, as described in Section~\ref{sec:method-extraction}. Additionally, both algorithms incorporate the same blind localization algorithm, as described in Section~\ref{sec:method-heatmap}.
The main difference between Comprint and Noiseprint is that Comprint exploits noise left by compression artifacts, whereas Noiseprint exploits noise unique to camera models. Since these are complementary strategies, we propose to fuse Comprint with Noiseprint. We define the fused method as \emph{Comprint+Noiseprint}. 

\begin{figure}[!t]
\includegraphics[width=\textwidth]{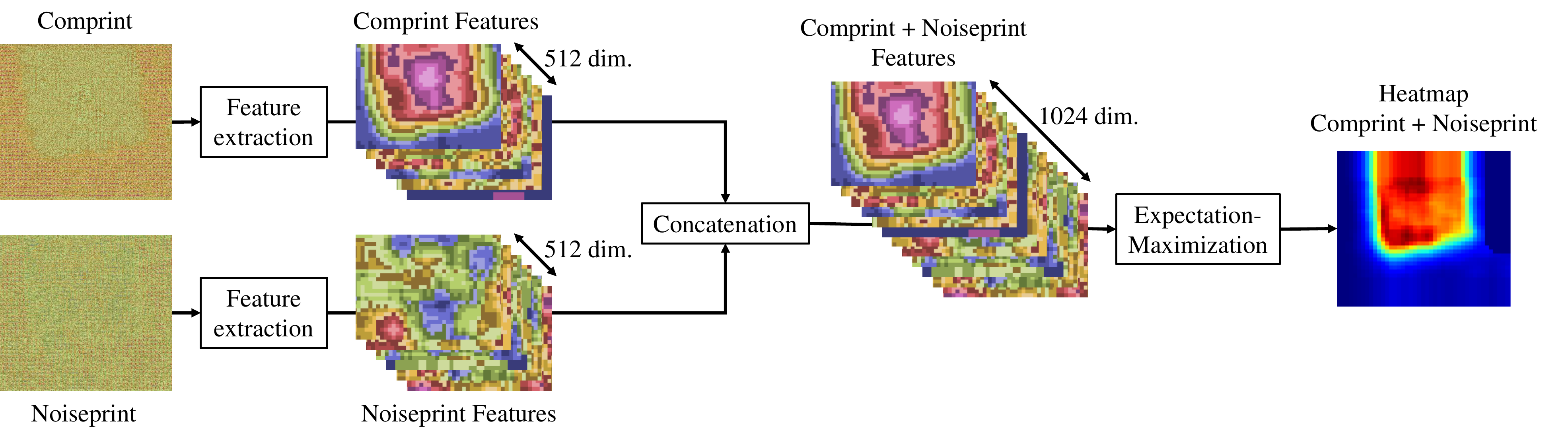}
\caption{Fusion diagram. The localization algorithm is adapted by concatenating the features of both Comprint and Noiseprint, and applying EM on the concatenated feature array.}
\label{fig:method-fusion-diagram}
\end{figure}

A high-level diagram of the fusion algorithm is given in Fig.~\ref{fig:method-fusion-diagram}, which adapts the localization process described in Section~\ref{sec:method-heatmap}. More specifically, the co-occurence-based features are additionally calculated from the noiseprint, instead of only from the comprint. Then, the multi-dimensional features from both fingerprints are stacked (in the third dimension). Subsequently, the stacked features are given to the EM algorithm that attempts to segment the features. In general, this stacking of fingerprint features can be used to apply fusion with any fingerprinting-based method. If the fused fingerprint contains complementary characteristics, we expect the performance to increase.

\newpage

\section{Evaluation}
\label{sec:evaluation}
This section evaluates the proposed Comprint method for forgery detection and localization, as well as the fusion between Comprint and Noiseprint. First, Section~\ref{sec:evaluation-setup} describes the experimental setup. Then, the results are presented and analysed in Section~\ref{sec:evaluation-results}.

\subsection{Experimental Setup}
\label{sec:evaluation-setup}

\subsubsection{Training Procedure}
\label{sec:evaluation-setup-training}
The training and validation images were obtained from the RAISE dataset~\cite{dangnguyen2015raise}. RAISE consists of high-resolution uncompressed images depicting various subjects and scenarios. From this dataset, 1000 images were randomly selected and used for training, and another 100 images were used for validation during training. The images were first converted to grayscale and resized to $200\times200$ pixels, similar to setup used for the DnCNN denoiser upon which our CNN is based~\cite{zhang2017beyond}. Using training images with a small resolution ensures that patches contain enough variation and information to properly train the network.

After loading the uncompressed images, we additionally apply on-the-fly cropping and JPEG compression during training. That is, for each input pair, the uncompressed input images are randomly cropped to an area of $48\times48$ pixels. The first patch of the pair is compressed with a randomly chosen JPEG Quality Factor (QF) or a randomly chosen quantization table, selected from a predefined list. The JPEG compression was done using python's Pillow library v9.2.0 and using Photoshop CS4 v11.0.
With probability $p=0.5$, the second patch of the pair is compressed using the same QF or quantization table. In the other case, it is compressed with another QF or quantization table from the predefined list.
Finally, both patches are subsequently fed to the CNN, in conjunction with the corresponding label that represents if they were compressed with the same (label $-1$) or a different QF/quantization table (label $+1$). In other words, in this paper, we limit the scope of the compression history by only utilizing a single compression step using the JPEG encoder with various QFs and quantization tables.

We used 10 different QFs as well as 9 commonly used quantization tables by Adobe Photoshop. That is, the QFs are chosen from the following predefined list: 20, 25, 30, 35, 40, 50, 60, 70, 80, 90, and the quantization tables correspond to Photoshop qualities 4 to 12.
For the QF list, a denser quality factor spacing is used at lower QFs because the difference between images compressed with slightly different quality factors is larger at lower QFs, due to the higher amount of noise being introduced.
Although only 19 different QFs and quantization tables are used, the training set is augmented to a very large size, because of the random crop and the random pairing of QFs or quantization tables.

The CNN was implemented with $d=20$ groups of convolutional layers.
Moreover, the denoising network that was used as initialization and the Siamese network were trained for 50 epochs, using 4000 batches per epoch, and 200 patch pairs per batch. Furthermore, we used the ADAM optimizer.
Future work will explore other, more advanced training setups.
For example, we could incorporate other image compression standards.

The code and data have been made available for reproducibility\footnote{Code available on \url{https://github.com/IDLabMedia/comprint}}.

\subsubsection{Reference methods}
\label{sec:evaluation-setup-ref-methods}
We compare the results of our proposed method to 13 reference methods that were proposed in the state of the art.
These can be classified in the four classes presented in Section~\ref{sec:sota}.
First, BLK~\cite{li2009blk}, DCT~\cite{ye2007dct}, NADQ~\cite{bianchi2012nadq}, ADQ~\cite{bianchi2011adq2}, and CAGI~\cite{iakovidou2018cagi} are conventional model-driven methods, and they all exploit JPEG artifacts. The implementations of these methods were provided by Iakovidou~\emph{et al.}~\cite{iakovidou2018cagi}.
Additionally, Splicebuster~\cite{cozzolino2015splicebuster} is also a model-driven method, but exploits traces left by in-camera processing rather than compression artifacts.
Second, DJPEG~\cite{park2018djpeg} and RRU-Net~\cite{bi2019rrunet} are supervised CNNs looking at specific clues (being double-JPEG-compression artifacts and splicing artifacts, respectively).
Third, ManTraNet~\cite{wu2019mantranet}, SPAN~\cite{xuefeng2020span}, and CR-CNN~\cite{yang2020crcnn} are generic supervised CNNs that are trained on a large variety of manipulations. Fourth and lastly, EXIF-SC~\cite{huh2018exifsc} and Noiseprint~\cite{cozzolino2020noiseprint} are one-class deep-learning methods trained on pristine data only, using Siamese networks.

\subsubsection{Evaluation Datasets}
\label{sec:evaluation-setup-datasets}
As we target our method to be applicable in the wild, we selected 9 evaluation datasets that cover a wide variety of manipulations (based on both editing software and AI). These are listed and described in Table~\ref{tab:evaluation-datasets}. 
The JPEG quality factors that are shown in the table are estimates that give an indication of the quality of the images in the datasets. That is because the QF is undefined when a non-standard quantization tables is used, and we consider the QF of the most similar standard quantization tables.

\begin{table}[!th]
\begin{center}
\caption{Datasets used for the forgery detection and localization evaluation.}
\label{tab:evaluation-datasets}
{
\renewcommand{\arraystretch}{1.3}
\footnotesize
\setlength\tabcolsep{4pt}%
\begin{tabular}{r@{\hskip4pt}lcccc}
\toprule
\multicolumn{2}{l}{Dataset} & \#fake & \#real & Format & Description\\
\midrule
\cite{fontani2013vipp} & VIPP & 62 & 69 & JPEG@40-100 &  Uses double JPEG compression\\
\cite{decarvalho2013dso1} & DSO-1 & 100 & 100 & PNG & Only splicing \\
\cite{zhou2017faceswap} & FaceSwap & 879 & 1651 & JPEG@39-100 & Face swaps with FaceSwap-app \\
\multirow{2}{*}[-2pt]{\cite{novozamsky2020imd2020}} & \multirow{2}{*}[-2pt]{IMD2020} & \multirow{2}{*}[-2pt]{2010} & \multirow{2}{*}[-2pt]{414} & PNG & Various forgery types found on \\
& & & & \& JPEG@45-100 & the internet, \emph{in the wild} \\
\cite{le2021openforensics} & OpenForensics & 18 895 & N/A & JPEG@100 & Synthetic face swapping \\
\bottomrule
\end{tabular}
}
\end{center}
\end{table}

\newpage

\subsubsection{Forgery Localization Measure}
\label{sec:evaluation-setup-measure-localization}
To perform binary pixel-level forgery localization, we can compare each pixel value of the continuous-valued heatmap to a certain threshold. This results in a binary heatmap that can be compared with the ground truth. For example, commonly-used measures such as the True Positive (TP) rate, False Positive (FP) rate and False Negative (FN) rate, precision, and recall can be calculated.
In this paper, we combine these measures in the F1 score, calculated as follows:

\begin{equation}
    F1 = \frac{1}{\dfrac{1}{\text{precision}}+\dfrac{1}{\text{recall}}} = \dfrac{2\text{TP}}{2\text{TP}+\text{FN}+\text{FP}}
\end{equation}
In other words, the F1 score is the harmonic mean of the precision and recall. In the equation, TP, FP and FN represents the number of pixel predictions that are True Positives, False Positives, and False Negatives, respectively.
Note that a higher F1 score (i.e., closer to 1) is better than a lower one (i.e., closer to 0).

Two remarks have to be made for this way of evaluating the localization performance.
First, to decouple the threshold value from the assessment, the maximum F1 score over all thresholds is taken.
Second, the proposed method makes a segmentation of regions that underwent different compression, and hence the choice whether a segment is forged or pristine is arbitrary. Therefore, both the regular and inverted ground truths are considered, and the maximum of the corresponding performance scores is kept. 

\subsubsection{Forgery Detection Measure}
\label{sec:evaluation-setup-measure-detection}
For image-level forgery detection evaluation, we first extract a global statistic from the heatmap. In this paper, we adopt the 99.5\textsuperscript{th} percentile highest value from the heatmap as a global statistic. Next, the global statistic can be compared to a threshold. In order to allow an evaluation that is independent of a certain threshold, we plot the TP rate against the FP rate for a range of thresholds, for each dataset. This plot is the Receiver Operating Characteristic (ROC) curve.
Finally, we calculate the Area Under the ROC Curve (AUC) as a performance measure. Note that a higher AUC score (i.e., closer to 1) is better than a lower one.

Note that other algorithms utilize other global statistics than the 99.5\textsuperscript{th} percentile highest value during image-level forgery detection. For example, ManTraNet~\cite{wu2019mantranet} and EXIF-SC~\cite{huh2018exifsc} utilize the average likelihood of the heatmap. However, the average will be significantly lower when the area of forgery is small. Additionally, DCT~\cite{ye2007dct} utilizes the maximum value, which solves the issue of small forgery regions. Nevertheless, the maximum value can be significantly affected by outliers. For this reason, we chose to utilize the 99.5\textsuperscript{th} percentile highest value as a global statistic instead.

In the experiments, we utilized the global statistic that was proposed in the original reference methods~\cite{wu2019mantranet, huh2018exifsc,ye2007dct}. When not clearly specified~\cite{bi2019rrunet}, we utilized the 99.5\textsuperscript{th} percentile highest value. Additionally, we did not calculate image-level forgery detection results when the reference methods only proposed pixel-level forgery localization~\cite{li2009blk, bianchi2012nadq, bianchi2011adq2, iakovidou2018cagi, park2018djpeg, xuefeng2020span, yang2020crcnn}. An exception for this are the Splicebuster~\cite{cozzolino2015splicebuster} and Noiseprint~\cite{cozzolino2020noiseprint} methods, for which we did perform image-level forgery detection using the 99.5\textsuperscript{th} percentile highest value. This was done because Comprint is based on these methods and hence an experimental comparison is essential.

\subsection{Results}
\label{sec:evaluation-results}
Table~\ref{tab:evaluation-results-f1} and Table~\ref{tab:evaluation-results-auc} summarize the F1 and AUC scores for forgery localization and detection, respectively. In the tables, the three largest values for each dataset are emphasized in bold, underline, and italics, respectively. Note that the AUC metric for forgery detection could not be calculated for OpenForensics since that dataset does not contain real images.
This section discusses several interesting observations from the presented results.

\begin{table}[!t]
\begin{center}%
\caption{Experimental pixel-level forgery localization results: average F1 score.}%
\label{tab:evaluation-results-f1}%
{%
\renewcommand{\arraystretch}{1.3}%
\footnotesize%
\setlength\tabcolsep{3pt}%
\begin{tabular}{r@{\hskip4pt}lcccccc}%
\toprule
\multicolumn{2}{l}{Model}	&	VIPP	&	DSO-1	&	FaceSwap	&	IMD2020	&	OpenFor	&	Average\\
\midrule
\cite{li2009blk} & BLK	&	0.411	&	0.449	&	0.097	&	0.267 	&	0.262 	&   0.297	\\
\cite{ye2007dct} & DCT	&	0.416	&	0.350	&	0.182	&	0.328 	&	0.408 	&   0.337	\\
\cite{bianchi2012nadq} & NADQ	&	0.248	&	0.247	&	0.037	&	0.170 	&	0.114 	&   0.163	\\
\cite{bianchi2011adq2} & ADQ	&	\underline{0.572}	&	0.530	&	\textbf{0.426}	&	\underline{0.446} 	&	\underline{0.675}	&   \textit{0.530}	\\
\cite{iakovidou2018cagi} & CAGI	&	0.460	&	0.537	&	0.172	&	0.299 	&	0.293	&   0.352	\\
\cite{cozzolino2015splicebuster} & Splicebuster	&	0.447	&	0.662	&	0.330	&	0.326	&	0.440	&	0.441\\[0.2cm]

\cite{park2018djpeg} & DJPEG	&	0.492	&	0.701	&	0.334	&	0.297	&	0.434	&	0.452\\
\cite{bi2019rrunet} & RRU-Net	&	0.349	&	0.360	&	0.064	&	0.336	&	0.203	&	0.262\\[0.2cm]

\cite{wu2019mantranet} & ManTraNet	&	0.354	&	0.538	&	0.153	&	0.324	&	0.653	&	0.404\\
\cite{xuefeng2020span} & SPAN	&	0.404	&	0.390 &	0.093	&	0.252	&	0.173	&	0.262\\
\cite{yang2020crcnn} & CR-CNN	&	0.406	&	0.432	&	0.128	&	\textbf{0.476}	&	0.224	&	0.333\\[0.2cm]

\cite{huh2018exifsc} & EXIF-SC	&	0.424	&	0.577	&	0.293	&	0.323	&	0.310	&	0.385\\
\cite{cozzolino2020noiseprint} & Noiseprint	&	\textit{0.556}	&	\underline{0.812}	&	0.309	&	0.397	&	\textit{0.672}	&	\underline{0.549}\\
\midrule
\multicolumn{2}{l}{Comprint}	&	0.497	&	\textit{0.763}	&	\textit{0.353}	&	0.388	&	0.634	&	0.527\\
\multicolumn{2}{l}{Comprint+Noiseprint}	&	\textbf{0.581}	&	\textbf{0.813}	&	\underline{0.417}	&	\textit{0.435}	&	\textbf{0.712}	&	\textbf{0.592}\\
\bottomrule
\end{tabular}
}
\end{center}
\end{table}

\paragraph{Comprint vs.~Noiseprint}
Comprint and Noiseprint often have similar scores. On average, Noiseprint performs slightly better than Comprint for forgery localization, but Comprint performs slightly better for forgery detection. Their resembling scores can be explained by the fact that they are based on the same architecture and use the same internal localization algorithm. However, it should be stressed that their underlying assumptions are completely different.


\begin{table}[!t]
\begin{center}
\caption{Experimental image-level forgery detection results: AUC score.}
\label{tab:evaluation-results-auc}
{
\renewcommand{\arraystretch}{1.3}
\footnotesize
\setlength\tabcolsep{3pt}%
\begin{tabular}{r@{\hskip4pt}lccccc}
\toprule
\multicolumn{2}{l}{Model}	&	VIPP	&	DSO-1	&	FaceSwap	&	IMD2020	&	Average\\
\midrule
\cite{ye2007dct} & DCT	        &	\textbf{0.653}	&	0.409	&	\textbf{0.569}	&	 \textit{0.600}	&   0.558	\\
\cite{cozzolino2015splicebuster} & Splicebuster	&	0.544	&	0.775	&	\underline{0.553}	&	0.544	&	0.604\\[0.2cm]

\cite{bi2019rrunet} & RRU-Net	    &	0.512	&	0.497	&	0.507	&	0.565	&	0.520\\[0.2cm]

\cite{wu2019mantranet} & ManTraNet	&	0.590	&	\textit{0.874}	&	0.479	&	\textit{0.558}	&	0.625\\[0.2cm]

\cite{huh2018exifsc} & EXIF-SC	&	0.397	&	0.237	&	0.453	&	0.435	&	0.381\\
\cite{cozzolino2020noiseprint} & Noiseprint	&	0.586	&	0.848	&	\textit{0.537}	&	0.548	&	\textit{0.630}\\
\midrule
\multicolumn{2}{l}{Comprint}	&	\underline{0.622}	&	\textbf{0.940}	&	0.525	&	\textbf{0.656}	&	\textbf{0.686}\\
\multicolumn{2}{l}{Comprint+Noiseprint}	&	\textit{0.616}	&	\underline{0.930}	&	0.532	&	\underline{0.637}	&	\underline{0.679}\\
\bottomrule
\end{tabular}
}
\end{center}
\end{table}

\paragraph{Comprint vs.~Splicebuster}
It is also interesting to compare Comprint with Splicebuster, as they use the same internal localization algorithm, and hence only differ in the input noise residual. In Splicebuster, this residual is obtained by high-pass filtering the input image, in contrast to Comprint which extracts a compression fingerprint. We can observe that, on average, Comprint performs significantly better than Splicebuster for both forgery detection and localization, highlighting its potential.

\clearpage

\paragraph{Double JPEG compression-based methods}
For the VIPP dataset, model-based methods using compression artifacts (such as DCT, ADQ, DJPEG and Comprint) are among the best performing methods.
This is not a coincidence, as this dataset was built with double compression in mind. 
In fact, for most datasets, the DCT method~\cite{ye2007dct} performs very well for forgery detection, and ADQ performs very well for forgery localization.
However, both methods have a significantly worse performance for DSO-1, which contains images in the losslessly compressed PNG format. In other words, DSO-1 may contain images that did not undergo double compression. This highlights that utilizing only double JPEG compression traces is not sufficient for forgery detection and localization.


\paragraph{Non-standard quantization tables}
For DSO-1, the model-based methods using compression artifacts perform significantly worse than the other methods. In contrast, the data-driven DJPEG-method which also used compression artifacts performs significantly better. This may be because it is trained using a large variety of non-standard quantization tables, in contrast to the model-based methods that only consider standard QFs. We noticed a similar worse performance when we trained Comprint using only standard QFs, i.e., without non-standard quantization tables. In that case, the F1 and AUC scores were only 0.526 and 0.759, respectively, in contrast to the much higher scores (0.763 and 0.94, respectively) that are established with the final proposed Comprint model that includes Photoshop quantization tables.
Future research can analyze more thoroughly how the usage of JPEG implementations or quantization tables unseen during training influences the performance.

\paragraph{Generalization to unseen forgeries}
The performance on FaceSwap and IMD2020 is relatively low for all methods (i.e., less than 0.5 F1 score), demonstrating that there is still a great need for more in-the-wild forgery localization research.
For IMD2020, CR-CNN has the best performance for forgery localization, yet it is among the worst performing methods for the other datasets, indicating its lack of generalization.
For OpenForensics, only Noiseprint, Comprint, ManTraNet and ADQ perform relatively well (F1 scores above 0.6). MantraNet exhibits a very poor performance on the other datasets, though. This again demonstrates the lack of generalization for supervised CNNs that include fake images in their dataset.
In contrast, Comprint and Noiseprint remain top of the class for all datasets, highlighting the superior generalization capabilities of one-class training methods.

\begin{figure*}
\centering
\subfloat[Forged image.]{\includegraphics[width=0.4\linewidth]{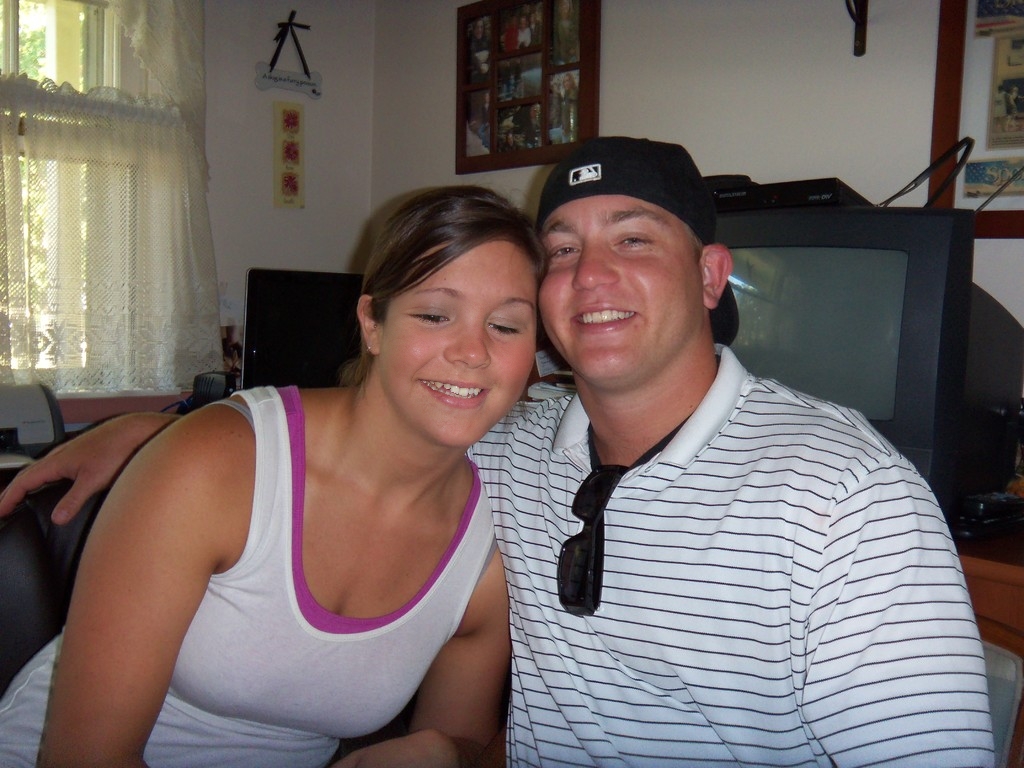}
\label{fig:evaluation-fusion-image}}
\hspace{1cm}
\subfloat[Ground truth.]{\includegraphics[width=0.4\linewidth]{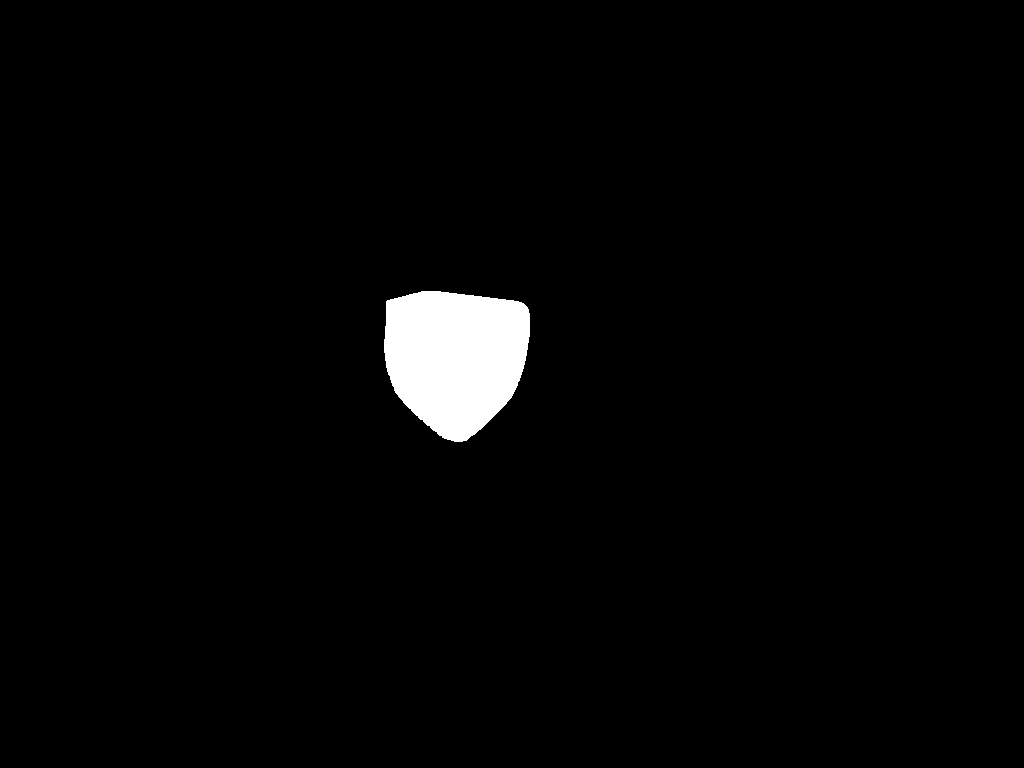}
\label{fig:evaluation-fusion-groundtruth}}

\subfloat[Comprint.]{\includegraphics[width=0.4\linewidth]{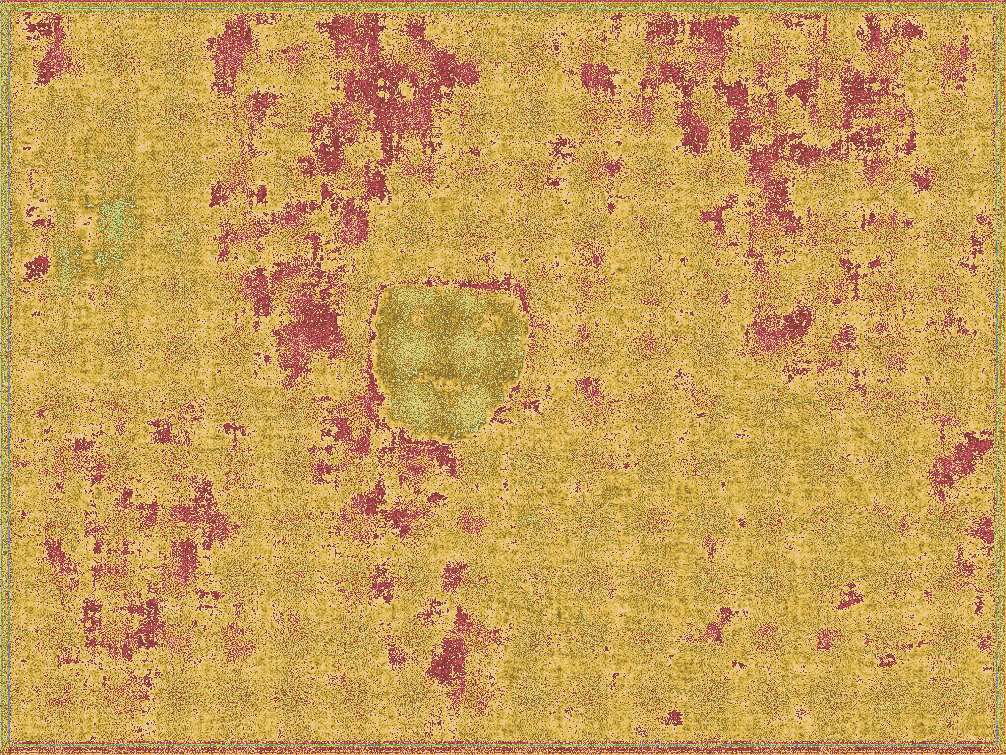}
\label{fig:evaluation-fusion-comprint}}
\hspace{1cm}
\subfloat[Noiseprint.]{\includegraphics[width=0.4\linewidth]{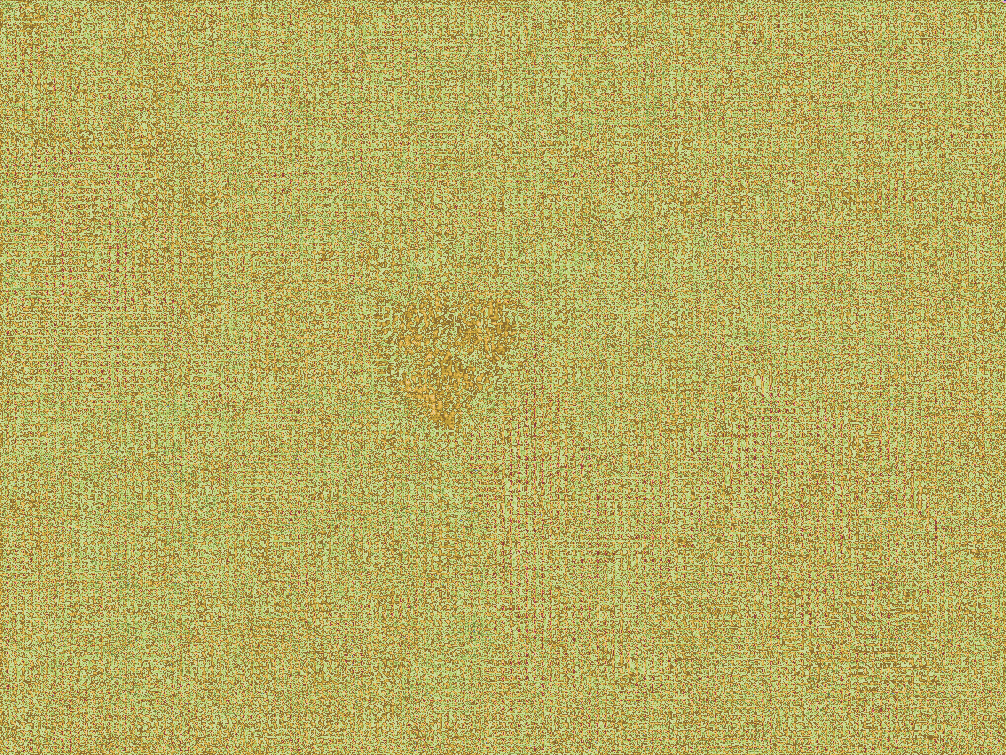}
\label{fig:evaluation-fusion-noiseprint}}

\subfloat[Comprint heatmap (F1 = 0.17).]{\includegraphics[width=0.4\linewidth]{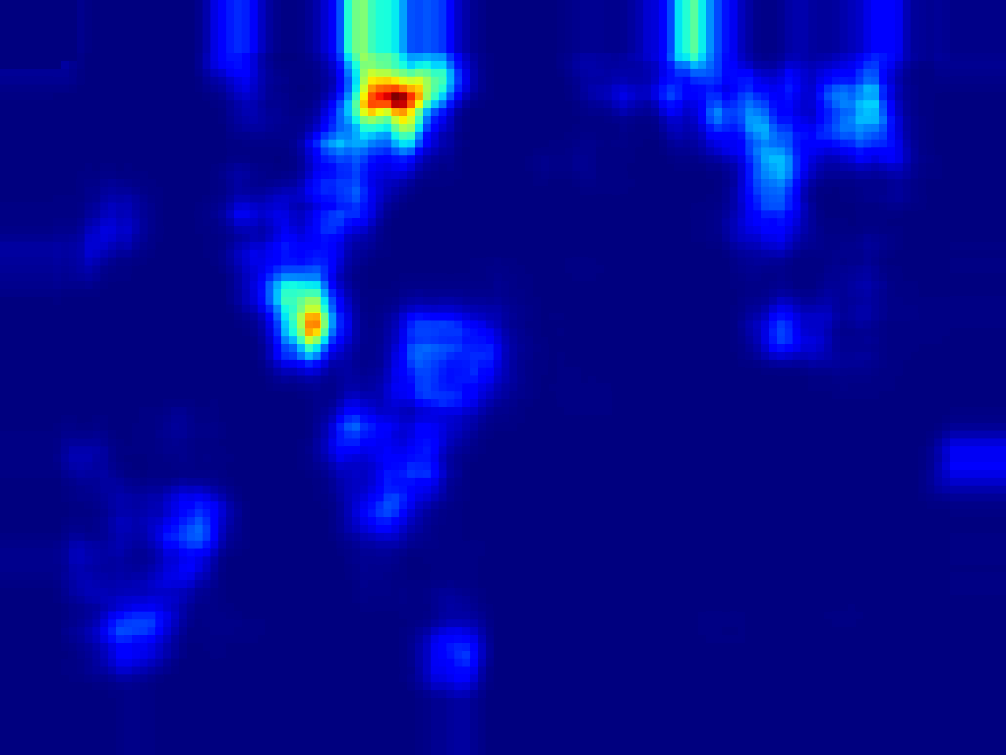}
\label{fig:evaluation-fusion-comprint-heatmap}}
\hspace{1cm}
\subfloat[Noiseprint heatmap (F1 = 0.43).]{\includegraphics[width=0.4\linewidth]{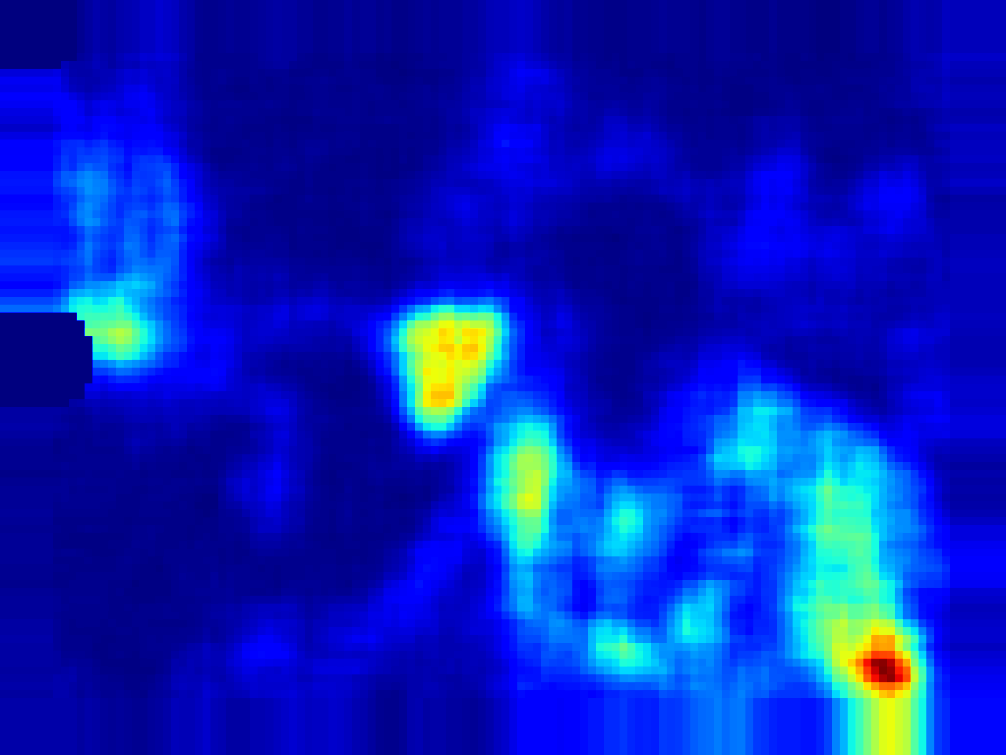}
\label{fig:evaluation-fusion-noiseprint-heatmap}}

\subfloat[Comprint+Noiseprint heatmap (F1 = 0.84).]{\includegraphics[width=0.4\linewidth]{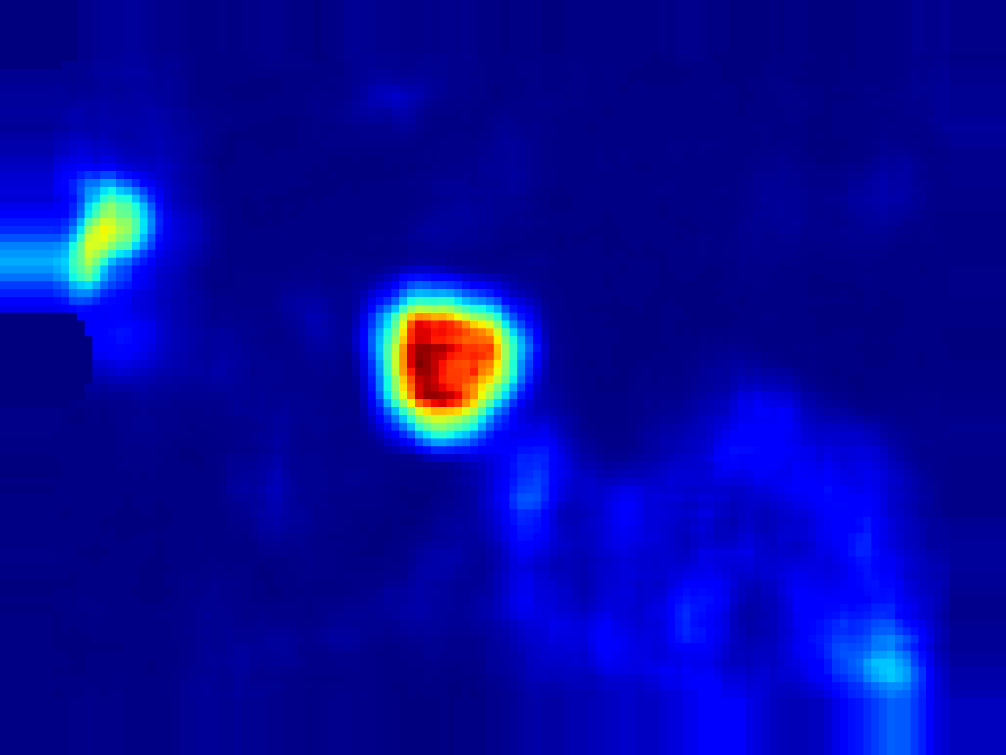}
\label{fig:evaluation-fusion-fused}}

\caption{Example of (a) image in which faces were manipulated (see (b) for ground truth). The fingerprints of (c) Comprint and (d) Noiseprint both generate (e, f) heatmaps that do not perform very well individually. In contrast, (g) the fusion Comprint+Noiseprint performs outstanding.
\label{fig:evaluation-fusion}}
\end{figure*}

\paragraph{Fusion: Comprint+Noiseprint}
For pixel-level forgery localization, the performance of both Comprint (F1 score of 0.527) and Noiseprint (F1 score of 0.549) improves when they are fused (F1 score of 0.592), on average.
For image-level forgery detection, Comprint performs slighlty better individually (AUC of 0.686) than the fusion (AUC of 0.679), on average.
Fig.~\ref{fig:evaluation-fusion} shows an example that highlights the positive effect of fusing Comprint and Noiseprint. Although the individual F1 scores are relatively low and their heatmaps contain false positive detections, the F1 score of the fusion is high and the corresponding heatmap is much more clear.

\paragraph{Top-performing methods}
For image-level forgery detection, Comprint (AUC of 0.686) outperforms all methods on average, although it is closely followed by Comprint+Noiseprint, and Noiseprint with average AUC scores of 0.685 and 0.630, respectively.
For pixel-level forgery localization, the fusion Comprint+Noiseprint (F1 score of 0.592) outperforms all evaluated methods, and is closely followed by Noiseprint (0.549), and Comprint (0.527).

\section{Conclusion}
\label{sec:conclusion}
This paper proposed Comprint, a forgery detection and localization method. The main novelty of our proposed method is the usage of the compression fingerprint that represents the compression history. The method is trained using only pristine data, i.e., images compressed in a different way. By detecting and localizing inconsistencies in the compression fingerprint, forgeries are exposed.
Additionally, we proposed to fuse Comprint with the complementary Noiseprint.

We demonstrated that Comprint and the fusion of Comprint and Noiseprint exhibit top-notch performance, unmatched by the 13 reference methods.
In general, in-the-wild forgery detection and localization is still challenging, though. Therefore, incorporating strategies for improved recompression robustness could be explored~\cite{wu2022robust}, as well as fusing more complementary methods. Additionally, Comprint's applicability on (deep)fake videos should be investigated as well.
In any case, in its current form, Comprint and the \emph{Comprint+Noiseprint} fusion can already be utilized to aid in-the-wild multimedia forensics.

\subsubsection{Acknowledgements}
This work was funded in part by the Research Foundation -- Flanders (FWO) under Grant V414022N, IDLab (Ghent University -- imec), Flanders Innovation \& Entrepreneurship (VLAIO), and the European Union. In addition, this material is based on research sponsored by the Defense Advanced Research Projects Agency (DARPA) and the Air Force Research Laboratory (AFRL) under agreement number FA8750-20-2-1004. 
The U.S. Government is authorized to reproduce and distribute reprints for Governmental purposes notwithstanding any copyright notation thereon. 
The views and conclusions contained herein are those of the authors and should not be interpreted as necessarily representing the official policies or endorsements, either expressed or implied, of DARPA and AFRL or the U.S. Government. This work is also supported by a Google gift and by the PREMIER project, funded by the Italian Ministry of Education, University, and Research within the PRIN 2017 program.

The computational resources (imec iLabt \& STEVIN Supercomputer Infrastructure) and services used in this work were kindly provided by Ghent University, imec, the Flemish Supercomputer Center (VSC), the Hercules Foundation, the Flemish Government department EWI, as well as by University Federico II of Naples.

\bibliographystyle{splncs04}
\bibliography{IEEEabrv, references-wo-doi}

\end{document}